\documentclass[11pt]{article}
\usepackage[margin=1in]{geometry}
\usepackage{iftex}
\ifXeTeX\usepackage{fontspec}\else\usepackage[utf8]{inputenc}\usepackage[T1]{fontenc}\usepackage{lmodern}\fi
\usepackage{microtype}
\usepackage{amsmath,amssymb}
\usepackage{graphicx}
\usepackage{booktabs}
\usepackage[small,labelfont=bf]{caption}
\usepackage[numbers,sort&compress]{natbib}
\usepackage{authblk}
\usepackage{xcolor}
\usepackage{xurl}
\usepackage[colorlinks,allcolors=blue!50!black]{hyperref}

\title{Scaling cross-situational word learning on children's naturalistic experiences}
\author[a]{Michael C. Frank\thanks{Corresponding author. E-mail: mcfrank@stanford.edu}}
\author[a]{Alvin W. M. Tan}
\author[b]{Khai Loong Aw}
\author[a,c]{Robert Z. Sparks}
\author[a]{Victor Moreno}
\author[d]{Grace Keene-Lee}
\author[e]{Meesha R. Ryan}
\author[a]{Tarun Sepuri}
\author[f]{Jane Yang}
\author[a]{Virginia A. Marchman}
\author[a,b]{Daniel L. K. Yamins}
\author[f]{Bria Long}
\affil[a]{Department of Psychology, Stanford University}
\affil[b]{Department of Computer Science, Stanford University}
\affil[c]{Department of Psychology, Harvard University}
\affil[d]{Department of Psychology, University of Wisconsin, Madison}
\affil[e]{Department of Psychology, University of Oregon}
\affil[f]{Department of Psychology, University of California, San Diego}
\date{}

\begin{document}
\maketitle

\begin{abstract}
How do children learn the meanings of the simple, concrete words that form a majority of their early vocabulary? Statistical information about many word meanings can be recovered from the gradual accumulation of information about the co-occurrence of words with their referents, a proposal known as ``cross-situational learning''. Contrastive vision--language models allow this learning strategy to be tested on egocentric video data capturing children's early real-life experiences with visual and linguistic inputs. Initial successes of this method on small datasets provide a proof-of-concept, but vision models struggle to learn robust visual representations when trained on these same egocentric videos. How dependent is cross-situational word learning performance on the quality of the model's visual representations? And can cross-situational models scale to larger and more diverse datasets? To address these questions, we trained models on the BabyView corpus, including 2,441 hours of longitudinal egocentric video from 48 children. Models trained on this dataset learned object--label mappings, but performance was far higher when they used internet-pretrained visual representations as opposed to when visual representations were learned solely from the developmental video data. Further, the majority of the learning signal was carried by a small proportion of utterances that described a visible referent. Models learned lexical representations that showed recognizable human-like semantic structure, but model performance showed substantially different trajectories over training experiences than those observed in children. Cross-situational learning is a viable word learning strategy for early noun vocabulary, but constructing developmentally plausible models will require understanding how to learn high-quality visual representations from naturalistic data.

\end{abstract}

Word learning is one of the most visible and important parts of children's development. While most children begin producing words around their first birthday, signs of comprehension can be detected even in infancy \cite{bergelson2012,tincoff1999,frank2021}. The earliest words include names for people (``mama'', ``dada''), for things (``ball'', ``shoe''), and even more abstract words (``no'', ``uh oh''). But as vocabulary expands to dozens or hundreds of words, across languages it tends to over-represent names for things -- nouns naming concrete objects \cite{frank2021,bates1994,hagihara2026}. How are these early noun labels learned? 

One influential proposal suggests that children learn through \emph{cross-situational word learning}: the gradual accumulation of statistical information about meaning through the co-occurrence of words and their referents \cite{pinker1984,yu2007rapid,smith2008infants}. On this account, children use their statistical learning abilities \cite{saffran1999statistical,aslin2017statistical} to track the objects that occur most frequently in the context of specific words, and consider these as possible candidate clues to reference -- and in turn, evidence for generalizations about meaning. 
For example, if a banana is often present when the word ``banana'' is uttered, the child might treat these observations as examples of a general category that could be a candidate meaning for the word.

The proposal that cross-situational learning is responsible for early lexical acquisition has been controversial \cite{gleitman2020easy}. Children likely require substantially more linguistic  scaffolding to learn verbs and other predicates compared with nouns, so cross-situational word learning will not work for all word types equally well \cite{gleitman1990,fisher1996structural}.  In addition, while adults \cite{yu2007rapid,kachergis2012associative,smith2011cross,roembke2023have} and children \cite{smith2008infants,yu2011you,vlach2013memory} can learn word meanings cross-situationally in laboratory experiments, evidence is more mixed on whether the same successes are observed in naturalistic data  \cite{medina2011words,yurovsky2013statistical} and whether, mechanistically, learners can simultaneously track multiple hypotheses \cite{trueswell2013,woodard2016two,yurovsky2015integrative,macdonald2017social}. Finally, there are theoretical bounds on the amount of information even a perfect learner could glean from an environment with high referential uncertainty \cite{blythe2010learning,vogt2012exploring}.

Egocentric videos -- recordings from the child's own perspective, typically made using head-mounted cameras -- have been a key piece of evidence in this debate: The natural statistics of the child's experience determine whether cross-situational learning is a viable strategy \cite{yu2021infant}. Indeed, experimental studies suggest that attentional and social cues constrain the  child's space of possible object referents, facilitating learning \cite{yurovsky2013statistical}, and objects in children's visual experience have a skewed distribution, in which a small number of frequent objects provide an early lever into the learning problem \cite{clerkin2017real,clerkin2022real,yang2026characterizingvisualrepresentationobjects}. Thus, a key test of the viability of cross-situational word learning is whether it can be applied to egocentric data from infants' real-life experiences. 

\begin{figure*}[t]
    \centering
    \includegraphics[width=\linewidth]{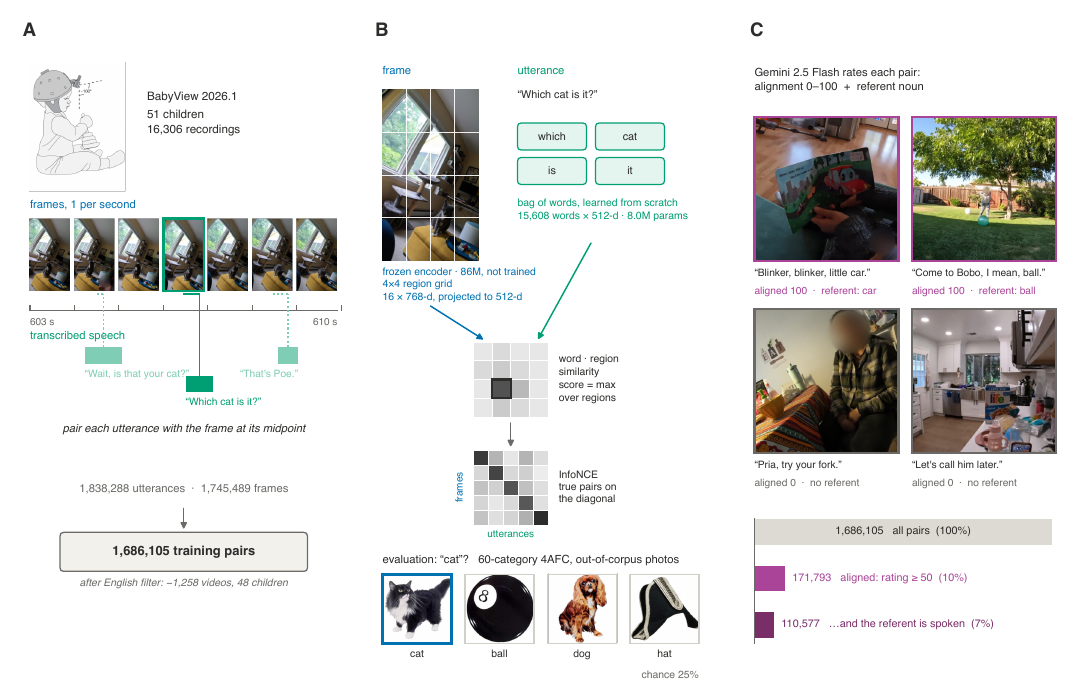}
    \caption{Our training pipeline. (A) The 2026.1 release of the BabyView corpus (N=51) was transcribed and downsampled to a set of frame--utterance pairs, then filtered on language (N=48). (B) Our VLM pipeline learned correspondences between the words in each utterance and the visual representation of subregions of the frame (as described by the visual encoding model). Primary evaluations were 4-alternative forced-choice trials constructed from an independent set of photos \cite{konkle2010conceptual}. (C) Each pair was automatically annotated for referential alignment (whether the utterance referred to something visible in the frame).} 
    \label{fig:pipeline}
\end{figure*}

Vision--language models (VLMs), neural networks that combine image data with text, provide a tool for testing claims about the viability of cross-situational learning. A rich earlier literature proposed specific computational instantiations of cross-situational learning strategies \cite{siskind1996computational,regier2005emergence,yu2007unified,frank2009using,fazly2010probabilistic,mcmurray2012,kachergis2012associative,stevens2017pursuit}, but this work was limited by the need to operate over schematic, symbolic data rather than actual images of the learning environment. 
In practice, the representation of a complex and messy situation might be rendered symbolically [[``look'', ``a'', ``ball''], [{\sc ball}, {\sc shoe}]] \cite{frank2009using}. 
Hence, a key critique of such models is that a child's visual environment is simply too cluttered or too complex to allow learning, and that the use of schematic data -- typically annotated by humans -- masked this fundamental issue \cite{medina2011words,gleitman2020easy}. Thus, here we focus on a simple VLM architecture that instantiates the principle of cross-situational learning \cite{vong2022cross}: contrastive image--text models \cite{radford2021learning}. These models learn visual and linguistic embeddings via a contrastive objective that constructs representations that are more similar to each other when linguistic and visual inputs co-occur than when they do not. 
This objective provides a computational instantiation of the core idea of cross-situational learning that can be applied to developmental egocentric video data.

Can such models learn from children's visual input? Using data derived from 60 hours of egocentric video from a single child \cite{sullivan2021saycam}, Vong et al. \cite{vong2024grounded} found robust evidence of word--object mapping in a contrastive VLM. The model was able to pair words with matching images using both labeled frames from the same child's experience and -- most impressively -- completely novel images from an object categorization dataset \cite{konkle2010conceptual}. A follow-up using data from two additional children suggested that these results could be replicated, though absolute accuracy was variable across datasets \cite{vong2026robustness}. These exciting results invite the question of whether this implementation of cross-situational learning could, in principle, learn a larger portion of a child's early noun vocabulary, given enough data. 

However, these results also do not separate the question of learning cross-situational mappings from its dependence on visual representation learning \cite{clerkin2022real}. Egocentric video, in particular, is challenging training data for visual representation learning \cite{orhan2024learning,long2025,lin2026egobabyvlm}. Thus, in our work here we decouple visual representation learning from learning cross-situational mappings. We pair a range of different pretrained vision encoders with learnable image--word mappings, asking what mappings can be learned across various visual representation conditions. This design thus provides a method for measuring how much a pure cross-situational learner -- as instantiated through a contrastive VLM -- could in principle learn from naturalistic data, as well as for determining the role of the quality of the visual representations in moderating this process.

In our experiments, we make use of the BabyView dataset \cite{long2025}, a growing dataset of egocentric videos from infants and children, leveraging the 2026.1 release (Figure \ref{fig:pipeline}A, Figures \ref{fig:corpus_effort}, \ref{fig:corpus_age}, \ref{fig:speech_density}). We first show that contrastive VLMs can achieve strong scaling, performing robustly on out-of-sample image--word pairings. However, scaling is far better when the visual representations come from an encoder trained on large amounts of internet data than when they come from an encoder trained on children's egocentric video data. Further, the majority of the informational value of the training data is carried by a small proportion of utterances that refer to an object that is present in the frame. The lexicons learned by the best models show semantic cluster structures, but model learning trajectories have a substantially different shape than children's. 

\section*{Results}

\subsection*{Word mapping performance depends on visual encoding}

\begin{figure*}[t]
    \centering
    \includegraphics[width=1\linewidth]{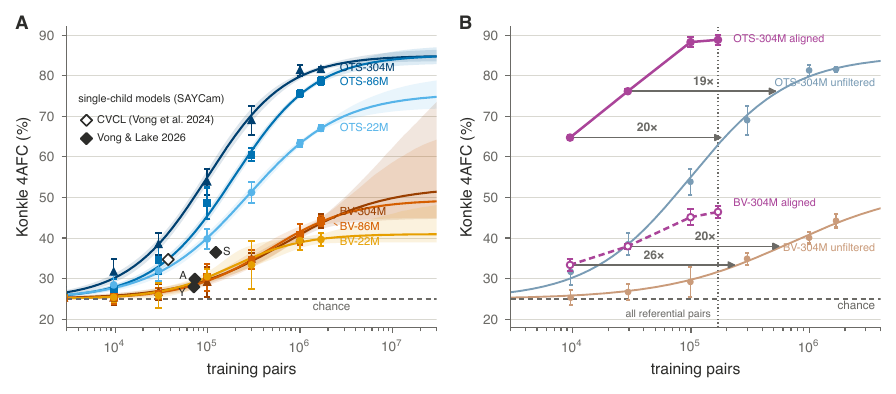}
    \caption{Model performance on the Konkle 4AFC evaluation, plotted by number of frame--utterance training pairs (log scale). (A) Performance split by vision encoder. OTS refers to models trained on internet-scale data, while BV refers to models trained on BabyView frames; numbers refer to model parameters. Points (mean $\pm$1 SD) are computed across model runs with different random seeds and subsamples of data. Sigmoid fits are 3 parameter logistic fits (lower asymptote fixed at 25\%), with 95\% confidence bands computed by Monte Carlo simulation. Black diamonds show data from Vong et al. (2024), open, and Vong \& Lake (2026), filled. (B) Scaling data for the OTS-304M and BV-304M models, trained only on filtered, referentially-aligned pairs. Blue and orange reproduce the unfiltered scaling curves from panel A. Arrows show how many times more unfiltered data are necessary to match the filtered model.}
    \label{fig:scaling}
\end{figure*}

For our primary experiments, we train the simple architecture pictured in Figure \ref{fig:pipeline}B, in which a pre-trained, frozen-weight vision encoder is paired with learned word embeddings via a single alignment layer. Models were trained on English-language data from the broader dataset (2,441 of 2,701 hours; 48 of 51 children). Visual representations were derived from vision transformer models trained using the DINOv3 objective \cite{simeoni2025dinov3}, a popular and well-performing self-supervised model for learning flexible visual representations. To evaluate learning with strong, general-purpose visual representations, we used ``off-the-shelf'' (OTS) models trained on approximately 1.7B curated, internet-sourced images, which yield high quality visual representations suitable for many downstream machine learning uses, and we examined small (22M parameters), base (86M), and large (304M) architectures \cite{simeoni2025dinov3}. We compared these to models pretrained from scratch on only the full, unfiltered set of BabyView frames (BV). 

To examine the developmental scaling of word learning performance, we trained multiple independent models at a range of scales (see Table \ref{tab:experiments} for a full list of all experiments). We evaluated models using the same metric as prior work, a 4-alternative-forced choice paradigm in which the model selected the image best matching a target word for a total of 60 common objects \cite{vong2024grounded}, using stock photos from a prior study \cite{konkle2010conceptual} (we refer to this as the ``Konkle'' evaluation). Each model's best training epoch was selected via testing on a held-out set of 117 additional object categories from the same set. Because the BabyView camera has a large vertical field of view \cite{long2024babyview}, we used multiple-instance learning (MIL) to select the best subregion of each frame to align to the language representation; in practice, this produced modest numerical increases, and only for some models and dataset sizes (Figure \ref{fig:nomil}).

Figure \ref{fig:scaling}A shows scaling performance across all vision encoders. The model using the largest off-the-shelf vision encoder showed the strongest performance, with a full dataset score of 81.6\% compared with chance (25\%). In contrast, models using BabyView trained encoders were robustly above chance for larger amounts of data, but showed far lower performance at all scales, with estimates matching prior work \cite{vong2026robustness}. The visual representations derived from all encoders were sufficient to categorize the Konkle test stimuli nearly perfectly (Figure \ref{fig:encoder_probe}), but they supported word mapping at very different levels when applied to children's complex and noisy visual input.

To ensure that the separation between BV and OTS vision encoders was not due to the domain-match between the Konkle evaluation and the OTS encoder (which both make use of third-person internet-derived photos), we created a within-domain word learning evaluation based on the BabyView data. Using held-out data from non-English videos, we constructed a set of in-domain 4AFC trials for 82 categories; these were sets of four frames where the object referred to by a target word was visible in only one. The OTS vision encoders still robustly out-performed the BV trained encoders, even though the domain advantage was reversed (Figure \ref{fig:indomain}) and there was only a modest advantage for the OTS models in visual prototype categorization (Figure \ref{fig:encoder_probe}). 

Data quantity was far more important than dataset diversity in determining performance: restricting training data to one or a small number of children did not meaningfully impact performance relative to a more diverse, quantity-matched sample (Figure \ref{fig:diversity}). Precise temporal alignment also did not meaningfully alter the pattern of results: MIL over a window of frames $\pm5$s away from the utterance produced some gains, primarily concentrated in the off-the-shelf models (Figure \ref{fig:window}).

\subsection*{Referential alignment is a key bottleneck}

Not all language that children hear is grounded in the physical context; research suggests that the most valuable data for early word--object mapping comes from a relatively small number of ``gems'' in which it is easy to ascertain reference \cite{bloom2002,cartmill2013quality, mollica2017,bergelson2017, gleitman2020easy,vong2024grounded}. To test this hypothesis, we used a commercial vision--language model (\texttt{gemini-2.5-flash}) to rate each frame--utterance pair on its referential alignment -- the degree to which the utterance referred to an object that was visible in the frame (Figure \ref{fig:pipeline}C). In total, 10\% of frames were judged to be referentially aligned (Figure \ref{fig:alignment_scores}), and 7\% contained a spoken referent; the rest typically included some form of deixis (e.g., ``look at that!''). These alignment ratings were related to human judgments in a stratified sample of 1500 frames (precision = 0.88 [95\% CI 0.85--0.90], recall = 0.84 [0.82--0.87], F1 = 0.86 [0.84--0.88]), suggesting a $\sim16\%$ rate of referential alignment in the full corpus. 

Training our best-performing model (OTS-304M) on only the referentially-aligned data produced results that were better than training on the entire corpus (88.8\% for the aligned subsample; Figure \ref{fig:scaling}B). The same result held true for the lower-performing BabyView trained visual encoders (full results across all encoders in Figure \ref{fig:alignment_encoders}). Filtering out non-referential utterances led to, if anything, \emph{better} performance compared with using the entire dataset. Critically, training on the full corpus with the aligned subsample removed dramatically reduced accuracy, while removing an equally-large unaligned subsample matched on utterance length and noun frequency modestly improved accuracy (Figure \ref{fig:alignment_controls}).

\subsection*{Learned representations align with human semantic structure}

\begin{figure*}[t]
    \centering
    \includegraphics[width=1\linewidth]{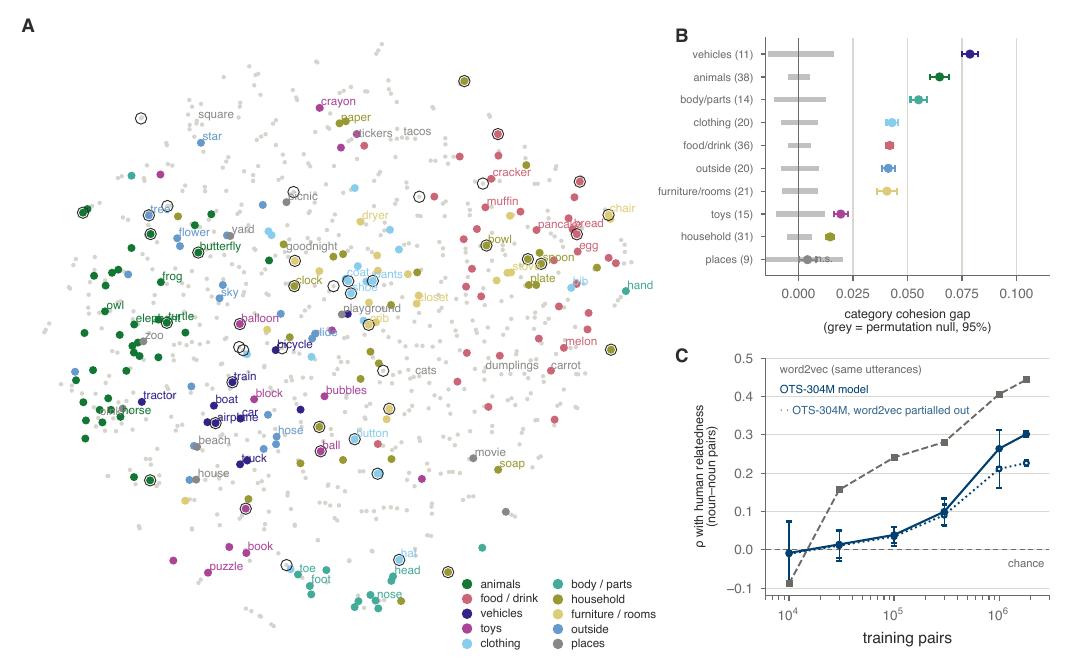}
    \caption{Assessing the lexicon learned by the top performing model (using the OTS-304M encoder). (A) t-distributed stochastic neighbor embedding (t-SNE) plot of lexical representations. Dots represent the set of 921 nouns whose top-3 nearest-neighbor similarity exceeds a threshold and that appeared 50 or more times in the training data. Object nouns appearing on the MB-CDI are colored by their semantic category on the instrument; words appearing on the Konkle evaluation are circled. (B) Mean category cohesion gap for each MB-CDI category: mean within-category cosine minus mean cosine to all other nouns. Gray bars show the 2.5\%-97.5\% band for a label-permuted null distribution. Error bars on points show between-seed SD. (C) Spearman correlation between model relatedness (cosine similarity) and human relatedness judgments for noun-noun pairs, plotted by training pairs (log scale). Correlations are computed across vocabulary present in the vocabulary at each scale. Lines show the OTS-304M model (solid) vs. a language-only word2vec model trained on the same data (dashed). Dotted line shows the partial correlation of the model, controlling for the word2vec correlation.}
    \label{fig:lexicon}
\end{figure*}

The models presented above use frozen visual representations and learn both vision--language mappings and lexical representations. How are these lexical representations structured? We examined whether the semantic structure of noun representations reflected the categories found in the MacArthur-Bates Communicative Development Inventories (MB-CDI), a standard instrument for measuring child language (Figure \ref{fig:lexicon}A). 

For the highest-performing visual encoder (OTS-304M), category label permutation suggested that these clusters for learned words were more coherent for 9 of 10 MB-CDI categories compared to chance (Figure \ref{fig:lexicon}B; see Figure \ref{fig:lexicon_tsne} for other encoders). The learned lexical representations for nouns in this OTS-304M model also showed modest but reliable correlations with human judgments. We used a composite word relatedness benchmark \cite{zhuang2024visual} to quantify the correlation between model similarities and human (adult) relatedness judgments across training scale (Figure \ref{fig:lexicon}C) and found a Spearman correlation of $\rho_{max}=.31$. Correlations were substantially lower across all words analyzed together (Figure \ref{fig:lexicon_relatedness}).

To compare this vision-derived score to one derived from linguistic structure, we trained word2vec models on the same utterances to capture word-to-word co-occurrence information \cite{mikolov2013efficient,hill2014simlex999evaluatingsemanticmodels}. Linguistic similarity correlations were higher than visually-driven similarities overall ($\rho_{max} = .45$, Figure \ref{fig:lexicon}C), but visual relatedness continued to be correlated with human judgments even when partialing out linguistic similarity. Intriguingly, models trained on referentially aligned pairs showed higher correlations with human relatedness judgments than would be expected given the total amount of data they were trained on, but lower than for the full dataset; thus, adding unaligned frames did increase similarity to human judgments (Figure \ref{fig:lexicon_alignment}).

\subsection*{Comparing model scaling with human learning trajectories}

\begin{figure*}[t]
    \centering
    \includegraphics[width=\linewidth]{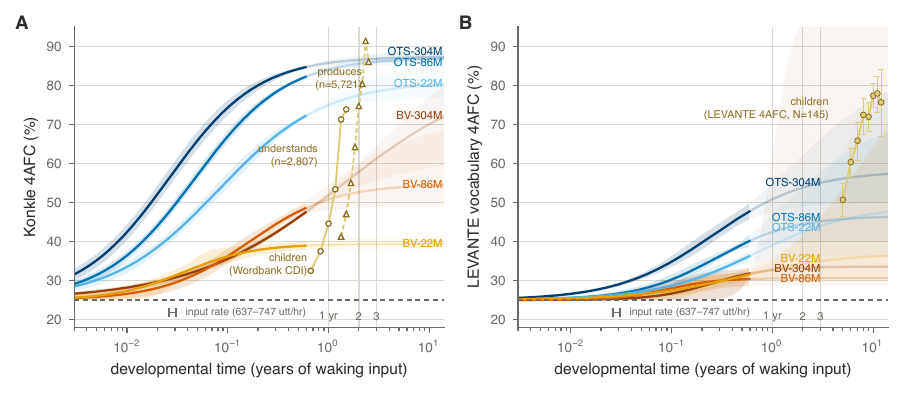}
    \caption{Scaling performance of models by developmental exposure time (years), assuming 4000 waking hours/year; faded lines indicate extrapolation. (A) Konkle evaluation fits and imputed proportion correct for children based on data from English-learning children in Wordbank. Evaluation is on the 40 words overlapping between the MB-CDI and the Konkle set. Error bar below the chance line is the 95\% confidence interval on the per-child median speech rate, indicating uncertainty on the input quantity to time mapping (horizontal) based on cross-child variation. (B) Model curves for the LEVANTE vocabulary evaluation, overlaid with LEVANTE children's proportion correct on these same words.}
    \label{fig:development}
\end{figure*}

In our final analysis, we compared the implied learning trajectories of our models to children's learning trajectories. There is no one accepted mapping from model training to developmental time, but as a first approximation, we rescaled the learning curves from Figure \ref{fig:scaling}A to developmental time, assuming a constant mapping of 4000 hours of learning input per year, and 697 utterances per hour (the average in our dataset). Since each frame--utterance pair is seen multiple times during training and pairs are presented in random order, these curves represent data-scaling functions, rather than literal developmental curves.  

We then used data from Wordbank \cite{frank2017}, a database of parent-reported vocabulary estimates for young children, to estimate children's accuracy on the same evaluation. While children recognize few of these words at one year of age -- despite having already experienced more hours of training data than our models -- their knowledge grows very rapidly across the second year (Figure \ref{fig:development}A). Individual word difficulties were modestly related at intermediate scales (Figure \ref{fig:item_difficulty}), but scaling was qualitatively different: initially delayed but ultimately far steeper for children than models. 

To investigate this trend using a more challenging test set, we turned to LEVANTE-bench \cite{tan2026levante}, which provides data from a forced-choice vocabulary measure in 5--12-year-old children \cite{kachergis2025creation} (Figure \ref{fig:development}B; see Figure \ref{fig:levante} for full LEVANTE scaling results). The OTS-304M model achieved a top performance of only 48\% on this challenging benchmark (reflecting in part that many test items did not occur in the training data). Again, scaling was much steeper in children than in models. 

\section*{Discussion}

Can a cross-situational learning strategy be applied at scale to naturalistic data to learn concrete nouns? Following previous work, we instantiated this strategy through contrastive VLMs. When trained on egocentric videos, these models found multi-modal correspondences that generalized to new stimuli with high accuracy. However, performance was much higher when models used strong visual representations derived from internet-scale training data than when they used visual representations derived from developmental data. Thus, the answer is a conditional yes: cross-situational learning can succeed, but only when it makes use of strong representations of the visual world. 

Visual representations serve both to handle the challenge posed by the complex and noisy visual environments that infants experience and to provide generalizable abstractions of object categories that allow learned labels to be generalized to new exemplars. Indeed, in our experiments, off-the-shelf visual representations were more useful even for in-domain evaluations based on egocentric data (Figures \ref{fig:encoder_probe} and \ref{fig:indomain}). The importance of visual representations for learning underscores classic arguments in the word learning literature about the importance of knowing that words name whole objects and that object categories are typically organized by shape \cite{markman1990constraints,smith2002}. These visual representations could be learned from data or (partially) inherited from evolved properties of the visual system \cite{aw2026zeroshotworldmodelsdevelopmentally}; they could be learned prior to, or jointly with, words. Regardless, they appear to be prerequisite for learning. 

Our findings also underscore the importance of referential alignment. Indeed, training a related class of models, Lin et al. \cite{lin2026egobabyvlm} found that alignment was a key factor controlling model performance in both naturalistic and artificial data. This finding is very consistent with the developmental literature: ``referential gems'' are thought to be quite rare in children's input \cite{gleitman2020easy,gillette1999human}, and children with more referentially-aligned input tend to learn faster than children who experience less referentially-aligned input \cite{cartmill2013quality}. In addition, toddlers are very sensitive to a speaker's referential focus \cite{baldwin1991infants}, and those toddlers who show stronger sensitivity to social cues -- which can be used to reveal alignment -- tend to learn words faster than children who show less sensitivity \cite{carpenter1998social,brooks2008infant}. One exciting possibility for future work would be to develop models that can bootstrap referential alignment, learning on the fly which utterances carry the most useful information \cite{frank2009using}.

Although contrastive models learned mappings effectively from a limited amount of data, they are a starting place, not a cognitive model of even concrete noun mapping. One signal of a mismatch between models and children is that our VLMs showed systematic differences from children in their scaling behavior. Models can begin learning mappings from even a small amount of data; in contrast, children show only traces of mapping in infancy \cite{bergelson2012,tincoff1999}, but they catch up very rapidly in their second year. We can only speculate about why this acceleration in learning might occur, but several -- mutually compatible -- possibilities include children's maturing memory and attentional abilities; their developing visual and conceptual representations; their increasing social inference abilities; and other developmental changes that might make their linguistic knowledge more visible to observers. Development of these abilities concurrently with word learning could lead to the kind of acceleration observed in children's learning trajectories \cite{frank2026children}; our study shows that grounding alone does not lead to this acceleration (cf. \cite{frank2023}). 

Our study is intended as a theoretical investigation of the potential viability of one version of the cross-situational learning strategy, instantiated through VLMs. We focused deliberately only on early noun learning, leaving other word classes for other models. The models we used are not intended as developmentally-realistic cognitive models: they use a bag-of-words representation with no syntactic structure and a frozen vision encoder, and they are trained via multiple passes through the same training data using gradient descent. Any correspondence between the models and human learning likely results from the computational structure of the problem rather than similarities in implementation between models and humans. Further, the generalizability of particular estimates is limited by the composition of the training data we used. The BabyView corpus is sampled to maximize data volume \cite{long2025}, rather than to represent a particular developmental population; many children are from high-income households and therefore, our sample consists of children from relatively homogeneous backgrounds. Due to privacy concerns, most of the data are recorded inside homes, rather than in public. Thus, an accurate characterization of children's visual environments would include data that are almost certainly more variable than those captured by our current dataset. 

Debates about the relevance of cross-situational learning to early vocabulary have not always clearly distinguished the twin challenges of visual representation learning and word mapping. Our work here suggests that research that examines one without the other will fall short of providing a complete picture of key developmental phenomena. We have shown that cross-situational word learning is a viable developmental mechanism, but building developmentally-appropriate models of visual representation learning and gaining more insights into how those models use referential alignment to learn word mappings remain critical next steps. We hope that continued advances in vision--language modeling lead to progress towards meeting these challenges.

\section*{Materials and Methods}

\subsection*{Data}

We used the 2026.1 release of the BabyView dataset, which contains 2,701 hours of data from N=51 children. Speech was transcribed using WhisperX (\texttt{large-v3})  \cite{bain2022whisperx, radfordRobustSpeechRecognition2022}. Because the BabyView dataset contains non-English data, we used \texttt{gemini-2.5-flash} for language identification of sampled 60-s windows. All language training excluded videos that had more than 50\% non-English data. This filter resulted in a final dataset of N=48 children contributing a total of 2,441 hours of video (min = 0.2, max = 192.1). Because speaker diarization is noisy, we chose to include all utterances (including the child's own speech) in the training data. Removing the child-labeled speech led to the same decrement in performance as removing random utterances (Figure \ref{fig:kchi}). 

Video was downsampled to 1 frame per second and each utterance was matched with the frame closest to its temporal midpoint (hence some frames were duplicated across several utterances). We used Gemini (\texttt{gemini-2.5-flash}) to annotate referential alignment on a 0--100 scale and we classified frames with scores $\geq50$ as aligned (Figure \ref{fig:alignment_scores}).  Children's vocabulary data were downloaded from Wordbank release 2026.1 using the \texttt{wordbankr} package and from LEVANTE via the \texttt{levante} package from \texttt{levante-data-pilots} version 3.0.

\subsection*{Neural network architecture}

Models paired a frozen image encoder and cached frame embeddings with trainable word representations (15,608 tokens x 512 dimensions) and a trainable alignment layer (LayerNorm, a linear projection to 512 dimensions, and a learned temperature). Frame embeddings were 16 encoder-dimensional sub-region embeddings (a 4$\times$4 average-pooled grid of patch features), with the maximally aligned region used in each training pass (via multiple instance learning). Training used an InfoNCE loss function over 20 epochs. Parameters were optimized using AdamW (learning rate 3e-4, batch size 256, weight decay 0.1). The epoch that resulted in the best fit to the held-out Konkle categories (N=117) was used for evaluation on the main set of categories (N=60). All models were trained on Nvidia A40 GPUs.

\section*{Data and code availability}
All code necessary to reproduce our work is available at \url{http://github.com/mcfrank/vlm-headcam}. BabyView videos and transcripts are available on \href{https://databrary.org}{Databrary.org}. BabyView-trained DINOv3 encoders are available on Hugging Face at \url{https://huggingface.co/collections/mcxfrank/babyview-dino-models}, with Databrary authorization.

\section*{Acknowledgments}
The authors gratefully acknowledge the families who participated in BabyView data collection. Funding for BabyView data collection was provided by the Stanford Human-Centered AI Initiative (HAI) Hoffman-Yee grant program, Schmidt Futures, Meta, Inc., The Stanford Center for the Study of Language and Information John Crosby Olney Fund, and NIH K99HD108386 to B.L.. Gemini annotations were supported by a gift from Google. Claude Code (Opus 4.8, 5, and 5.5 and Fable 5.0 and 5.1) was used for data analysis, model development, model training, visualization, and reproducibility checks.

\section*{Author contributions}
M.C.F., A.W.M.T., K.L.A., D.L.K.Y., and B.L. designed research; M.C.F., R.Z.S., V.A.Mo., G.K.-L., M.R.R., T.S., J.Y., and V.A.Ma. performed research; M.C.F., A.W.M.T., and K.L.A. contributed new analytic tools; M.C.F., A.W.M.T., K.L.A., T.S., and J.Y. analyzed data; and M.C.F., A.W.M.T., R.Z.S., V.A.Ma., and B.L. wrote the paper.

\section*{Competing interests}
The authors declare no conflicts of interest.

\bibliographystyle{unsrtnat}
\bibliography{vlm-headcam}

@misc{aw2026zeroshotworldmodelsdevelopmentally,
      title={Zero-shot World Models Are Developmentally Efficient Learners}, 
      author={Khai Loong Aw and Klemen Kotar and Wanhee Lee and Seungwoo Kim and Khaled Jedoui and Rahul Venkatesh and Lilian Naing Chen and Michael C. Frank and Daniel L. K. Yamins},
      year={2026},
      eprint={2604.10333},
      archivePrefix={arXiv},
      primaryClass={cs.AI},
      url={https://arxiv.org/abs/2604.10333}, 
}

@article{regier2005emergence,
  title={The emergence of words: Attentional learning in form and meaning},
  author={Regier, Terry},
  journal={Cognitive science},
  volume={29},
  number={6},
  pages={819--865},
  year={2005},
  publisher={Wiley Online Library}
}

@article{siskind1996computational,
  title={A computational study of cross-situational techniques for learning word-to-meaning mappings},
  author={Siskind, Jeffrey Mark},
  journal={Cognition},
  volume={61},
  number={1-2},
  pages={39--91},
  year={1996},
  publisher={Elsevier}
}

@article{frank2009using,
  title={Using speakers' referential intentions to model early cross-situational word learning},
  author={Frank, Michael C and Goodman, Noah D and Tenenbaum, Joshua B},
  journal={Psychological science},
  volume={20},
  number={5},
  pages={578--585},
  year={2009},
  publisher={SAGE Publications Sage CA: Los Angeles, CA}
}

@article{yu2007unified,
  title={A unified model of early word learning: Integrating statistical and social cues},
  author={Yu, Chen and Ballard, Dana H},
  journal={Neurocomputing},
  volume={70},
  number={13-15},
  pages={2149--2165},
  year={2007},
  publisher={Elsevier}
}

@article{kachergis2012associative,
  title={An associative model of adaptive inference for learning word--referent mappings},
  author={Kachergis, George and Yu, Chen and Shiffrin, Richard M},
  journal={Psychonomic bulletin \& review},
  volume={19},
  number={2},
  pages={317--324},
  year={2012},
  publisher={Springer}
}

@article{stevens2017pursuit,
  title={The pursuit of word meanings},
  author={Stevens, Jon Scott and Gleitman, Lila R and Trueswell, John C and Yang, Charles},
  journal={Cognitive science},
  volume={41},
  pages={638--676},
  year={2017},
  publisher={Wiley Online Library}
}

@article{fazly2010probabilistic,
  title={A probabilistic computational model of cross-situational word learning},
  author={Fazly, Afsaneh and Alishahi, Afra and Stevenson, Suzanne},
  journal={Cognitive Science},
  volume={34},
  number={6},
  pages={1017--1063},
  year={2010},
  publisher={Wiley Online Library}
}

@article{clerkin2017real,
  title={Real-world visual statistics and infants' first-learned object names},
  author={Clerkin, Elizabeth M and Hart, Elizabeth and Rehg, James M and Yu, Chen and Smith, Linda B},
  journal={Philosophical Transactions of the Royal Society B: Biological Sciences},
  volume={372},
  number={1711},
  pages={20160055},
  year={2017}
}

@article{clerkin2022real,
  title={Real-world statistics at two timescales and a mechanism for infant learning of object names},
  author={Clerkin, Elizabeth M and Smith, Linda B},
  journal={Proceedings of the National Academy of Sciences},
  volume={119},
  number={18},
  pages={e2123239119},
  year={2022},
  publisher={National Academy of Sciences}
}

@article{yu2021infant,
  title={The infant’s view redefines the problem of referential uncertainty in early word learning},
  author={Yu, Chen and Zhang, Yayun and Slone, Lauren K and Smith, Linda B},
  journal={Proceedings of the National Academy of Sciences},
  volume={118},
  number={52},
  pages={e2107019118},
  year={2021},
  publisher={National Academy of Sciences}
}

@article{vong2022cross,
  title={Cross-situational word learning with multimodal neural networks},
  author={Vong, Wai Keen and Lake, Brenden M},
  journal={Cognitive science},
  volume={46},
  number={4},
  pages={e13122},
  year={2022},
  publisher={Wiley Online Library}
}

@article{vong2026robustness,
  title={On the robustness of modeling grounded word learning through a child’s egocentric input},
  author={Vong, Wai Keen and Lake, Brenden M},
  journal={Open Mind},
  volume={10},
  pages={1253--1278},
  year={2026},
  publisher={MIT Press}
}

@article{konkle2010conceptual,
  title={Conceptual distinctiveness supports detailed visual long-term memory for real-world objects.},
  author={Konkle, Talia and Brady, Timothy F and Alvarez, George A and Oliva, Aude},
  journal={Journal of experimental Psychology: general},
  volume={139},
  number={3},
  pages={558},
  year={2010},
  publisher={American Psychological Association}
}

@article{cartmill2013quality,
  title={Quality of early parent input predicts child vocabulary 3 years later},
  author={Cartmill, Erica A and Armstrong III, Benjamin F and Gleitman, Lila R and Goldin-Meadow, Susan and Medina, Tamara N and Trueswell, John C},
  journal={Proceedings of the National Academy of Sciences},
  volume={110},
  number={28},
  pages={11278--11283},
  year={2013},
  publisher={National Academy of Sciences}
}

@article{long2024babyview,
  title={The BabyView camera: designing a new head-mounted camera to capture children’s early social and visual environments},
  author={Long, Bria and Goodin, Sarah and Kachergis, George and Marchman, Virginia A and Radwan, Samaher F and Sparks, Robert Z and Xiang, Violet and Zhuang, Chengxu and Hsu, Oliver and Newman, Brett and others},
  journal={Behavior Research Methods},
  volume={56},
  number={4},
  pages={3523--3534},
  year={2024},
  publisher={Springer}
}

@article{simeoni2025dinov3,
  title={Dinov3},
  author={Sim{\'e}oni, Oriane and Vo, Huy V and Seitzer, Maximilian and Baldassarre, Federico and Oquab, Maxime and Jose, Cijo and Khalidov, Vasil and Szafraniec, Marc and Yi, Seungeun and Ramamonjisoa, Micha{\"e}l and others},
  journal={arXiv preprint arXiv:2508.10104},
  year={2025}
}

@article{vong2024grounded,
  title={Grounded language acquisition through the eyes and ears of a single child},
  author={Vong, Wai Keen and Wang, Wentao and Orhan, A Emin and Lake, Brenden M},
  journal={Science},
  volume={383},
  number={6682},
  pages={504--511},
  year={2024},
  publisher={American Association for the Advancement of Science}
}

@article{sullivan2021saycam,
  title={SAYCam: A large, longitudinal audiovisual dataset recorded from the infant’s perspective},
  author={Sullivan, Jessica and Mei, Michelle and Perfors, Andrew and Wojcik, Erica and Frank, Michael C},
  journal={Open mind},
  volume={5},
  pages={20--29},
  year={2021},
  publisher={MIT Press One Rogers Street, Cambridge, MA 02142-1209, USA journals-info~…}
}

@misc{yang2026characterizingvisualrepresentationobjects,
      title={Characterizing the visual representation of objects from the child's view}, 
      author={Jane Yang and Tarun Sepuri and Alvin Wei Ming Tan and Khai Loong Aw and Michael C. Frank and Bria Long},
      year={2026},
      eprint={2605.14990},
      archivePrefix={arXiv},
      primaryClass={cs.CV},
      url={https://arxiv.org/abs/2605.14990}, 
}

@inproceedings{radford2021learning,
  title={Learning transferable visual models from natural language supervision},
  author={Radford, Alec and Kim, Jong Wook and Hallacy, Chris and Ramesh, Aditya and Goh, Gabriel and Agarwal, Sandhini and Sastry, Girish and Askell, Amanda and Mishkin, Pamela and Clark, Jack and others},
  booktitle={International conference on machine learning},
  pages={8748--8763},
  year={2021},
  organization={PmLR}
}

@article{medina2011words,
  title={How words can and cannot be learned by observation},
  author={Medina, Tamara Nicol and Snedeker, Jesse and Trueswell, John C and Gleitman, Lila R},
  journal={Proceedings of the National Academy of Sciences},
  volume={108},
  number={22},
  pages={9014--9019},
  year={2011},
  publisher={National Academy of Sciences}
}

@article{yurovsky2015integrative,
  title={An integrative account of constraints on cross-situational learning},
  author={Yurovsky, Daniel and Frank, Michael C},
  journal={Cognition},
  volume={145},
  pages={53--62},
  year={2015},
  publisher={Elsevier}
}

@article{yurovsky2013statistical,
  title={Statistical word learning at scale: The baby's view is better},
  author={Yurovsky, Daniel and Smith, Linda B and Yu, Chen},
  journal={Developmental science},
  volume={16},
  number={6},
  pages={959--966},
  year={2013},
  publisher={Wiley Online Library}
}

@article{smith2011cross,
  title={Cross-situational learning: An experimental study of word-learning mechanisms},
  author={Smith, Kenny and Smith, Andrew DM and Blythe, Richard A},
  journal={Cognitive Science},
  volume={35},
  number={3},
  pages={480--498},
  year={2011},
  publisher={Wiley Online Library}
}

@article{vogt2012exploring,
  title={Exploring the robustness of cross-situational learning under Zipfian distributions},
  author={Vogt, Paul},
  journal={Cognitive Science},
  volume={36},
  number={4},
  pages={726--739},
  year={2012},
  publisher={Wiley Online Library}
}

@article{blythe2010learning,
  title={Learning times for large lexicons through cross-situational learning},
  author={Blythe, Richard A and Smith, Kenny and Smith, Andrew DM},
  journal={Cognitive Science},
  volume={34},
  number={4},
  pages={620--642},
  year={2010},
  publisher={Wiley Online Library}
}

@article{fisher1996structural,
  title={Structural limits on verb mapping: The role of analogy in children's interpretations of sentences},
  author={Fisher, Cynthia},
  journal={Cognitive psychology},
  volume={31},
  number={1},
  pages={41--81},
  year={1996},
  publisher={Elsevier}
}

@article{woodard2016two,
  title={Two-and three-year-olds track a single meaning during word learning: Evidence for propose-but-verify},
  author={Woodard, Kristina and Gleitman, Lila R and Trueswell, John C},
  journal={Language learning and development},
  volume={12},
  number={3},
  pages={252--261},
  year={2016},
  publisher={Taylor \& Francis}
}

@article{macdonald2017social,
  title={Social cues modulate the representations underlying cross-situational learning},
  author={MacDonald, Kyle and Yurovsky, Daniel and Frank, Michael C},
  journal={Cognitive psychology},
  volume={94},
  pages={67--84},
  year={2017},
  publisher={Elsevier}
}

@article{yu2011you,
  title={What you learn is what you see: using eye movements to study infant cross-situational word learning},
  author={Yu, Chen and Smith, Linda B},
  journal={Developmental science},
  volume={14},
  number={2},
  pages={165--180},
  year={2011},
  publisher={Wiley Online Library}
}

@article{vlach2013memory,
  title={Memory constraints on infants’ cross-situational statistical learning},
  author={Vlach, Haley A and Johnson, Scott P},
  journal={Cognition},
  volume={127},
  number={3},
  pages={375--382},
  year={2013},
  publisher={Elsevier}
}

@article{roembke2023have,
  title={What have we learned from 15 years of research on cross-situational word learning? A focused review},
  author={Roembke, Tanja C and Simonetti, Matilde E and Koch, Iring and Philipp, Andrea M},
  journal={Frontiers in Psychology},
  volume={14},
  pages={1175272},
  year={2023},
  publisher={Frontiers Media SA}
}

@article{saffran1999statistical,
  title={Statistical learning of tone sequences by human infants and adults},
  author={Saffran, Jenny R and Johnson, Elizabeth K and Aslin, Richard N and Newport, Elissa L},
  journal={Cognition},
  volume={70},
  number={1},
  pages={27--52},
  year={1999},
  publisher={Elsevier}
}

@article{aslin2017statistical,
  title={Statistical learning: A powerful mechanism that operates by mere exposure},
  author={Aslin, Richard N},
  journal={Wiley Interdisciplinary Reviews: Cognitive Science},
  volume={8},
  number={1-2},
  pages={e1373},
  year={2017},
  publisher={Wiley Online Library}
}

@article{smith2008infants,
  title={Infants rapidly learn word-referent mappings via cross-situational statistics},
  author={Smith, Linda and Yu, Chen},
  journal={Cognition},
  volume={106},
  number={3},
  pages={1558--1568},
  year={2008},
  publisher={Elsevier}
}

@article{yu2007rapid,
  title={Rapid word learning under uncertainty via cross-situational statistics},
  author={Yu, Chen and Smith, Linda B},
  journal={Psychological science},
  volume={18},
  number={5},
  pages={414--420},
  year={2007},
  publisher={SAGE Publications Sage CA: Los Angeles, CA}
}

@book{pinker1984,
  title={Language learnability and language development},
  author={Pinker, Steven},
  year={1984},
  publisher={Harvard University Press}
}

@inproceedings{hagihara2026,
  title={Lexical category biases in children's early vocabulary: Decomposing linguistic and cultural predictors across 32 languages and dialects},
  author={Hagihara, Hiromichi and Tan, Alvin Wei Ming and Frank, Michael C},
  booktitle={Topics in Cognitive Science},
  year={in press}
}

@article{tincoff1999,
  title={Some beginnings of word comprehension in 6-month-olds},
  author={Tincoff, Ruth and Jusczyk, Peter W},
  journal={Psychological science},
  volume={10},
  number={2},
  pages={172--175},
  year={1999},
  publisher={SAGE Publications Sage CA: Los Angeles, CA}
}

@article{gleitman1990,
  title={The structural sources of verb meanings},
  author={Gleitman, Lila},
  journal={Language acquisition},
  volume={1},
  number={1},
  pages={3--55},
  year={1990},
  publisher={Taylor \& Francis}
}

@article{bates1994,
  title={Developmental and stylistic variation in the composition of early vocabulary},
  author={Bates, Elizabeth and Marchman, Virginia and Thal, Donna and Fenson, Larry and Dale, Philip and Reznick, J Steven and Reilly, Judy and Hartung, Jeff},
  journal={Journal of child language},
  volume={21},
  number={1},
  pages={85--123},
  year={1994},
  publisher={Cambridge University Press}
}

@article{mcmurray2012,
  title={Word learning emerges from the interaction of online referent selection and slow associative learning.},
  author={McMurray, Bob and Horst, Jessica S and Samuelson, Larissa K},
  journal={Psychological review},
  volume={119},
  number={4},
  pages={831},
  year={2012},
  publisher={American Psychological Association}
}

@inproceedings{zhuang2024visual,
  title={Visual grounding helps learn word meanings in low-data regimes},
  author={Zhuang, Chengxu and Fedorenko, Evelina and Andreas, Jacob},
  booktitle={Proceedings of the 2024 Conference of the North American Chapter of the Association for Computational Linguistics: Human Language Technologies (Volume 1: Long Papers)},
  pages={1311--1329},
  year={2024}
}

@article{orhan2024learning,
  title={Learning high-level visual representations from a child’s perspective without strong inductive biases},
  author={Orhan, A Emin and Lake, Brenden M},
  journal={Nature Machine Intelligence},
  volume={6},
  number={3},
  pages={271--283},
  year={2024},
  publisher={Nature Publishing Group UK London}
}

@article{tan2026levante,
  title={LEVANTE-bench: Multi-Scale Comparison of VLMs to Children Using Cognitive Tasks (or, "Is Your VLM Smarter Than a 5th Grader?")},
  author={Tan, Alvin Wei Ming and Cardinal, David and Lorido-Botran, Tania and Bravo-Sanchez, Laura and Yu, Sunny and Frank, Michael C},
  journal={arXiv preprint arXiv:2606.05497},
  year={2026}
}

@article{frank2026children,
  title={Children, but not language models, show accelerating returns in word learning},
  author={Frank, Michael C},
  journal={arXiv preprint arXiv:2608.17120},
  year={2026}
}

@article{brooks2008infant,
  title={Infant gaze following and pointing predict accelerated vocabulary growth through two years of age: A longitudinal, growth curve modeling study},
  author={Brooks, Rechele and Meltzoff, Andrew N},
  journal={Journal of child language},
  volume={35},
  number={1},
  pages={207--220},
  year={2008},
  publisher={Cambridge University Press}
}

@article{carpenter1998social,
  title={Social cognition, joint attention, and communicative competence from 9 to 15 months of age},
  author={Carpenter, Malinda and Nagell, Katherine and Tomasello, Michael and Butterworth, George and Moore, Chris},
  journal={Monographs of the society for research in child development},
  pages={i--174},
  year={1998},
  publisher={JSTOR}
}

@article{gillette1999human,
  title={Human simulations of vocabulary learning},
  author={Gillette, Jane and Gleitman, Henry and Gleitman, Lila and Lederer, Anne},
  journal={Cognition},
  volume={73},
  number={2},
  pages={135--176},
  year={1999},
  publisher={Elsevier}
}

@article{gleitman2020easy,
  title={Easy words: Reference resolution in a malevolent referent world},
  author={Gleitman, Lila R and Trueswell, John C},
  journal={Topics in cognitive science},
  volume={12},
  number={1},
  pages={22--47},
  year={2020},
  publisher={Wiley Online Library}
}

@article{baldwin1991infants,
  title={Infants' contribution to the achievement of joint reference},
  author={Baldwin, Dare A},
  journal={Child development},
  volume={62},
  number={5},
  pages={875--890},
  year={1991},
  publisher={Oxford University Press}
}

@article{lin2026egobabyvlm,
  title={EgoBabyVLM: Benchmarking Cross-Modal Learning from Naturalistic Egocentric Video Data},
  author={Lin, Dongyan and Rust, Phillip and Corrales, Angel Villar and Tan, Alvin WM and Luthra, Mahi and Saint-James, Charles-{\'E}ric and Moritz, Rashel and Krogh-Jespersen, Sheila and Stark, Vanessa and Parimi, Surya and others},
  journal={arXiv preprint arXiv:2605.19130},
  year={2026}
}

@article{markman1990constraints,
  title={Constraints children place on word meanings},
  author={Markman, Ellen M},
  journal={Cognitive science},
  volume={14},
  number={1},
  pages={57--77},
  year={1990},
  publisher={Wiley Online Library}
}

@misc{kachergis2025creation,
  title={Creation and validation of the LEVANTE core tasks: Internationalized measures of learning and development for children ages 5-12 years},
  author={Kachergis, George and O'Reilly, Fionnuala and Braginsky, Mika and Xiao, Xingyao and Lightbody, Amy and Shannon, KA and Watson, Zachary and Zhang, Lijin and Zhu, Rebecca and Abutto, AB and others},
  year={2025}
}

@misc{hill2014simlex999evaluatingsemanticmodels,
      title={SimLex-999: Evaluating Semantic Models with (Genuine) Similarity Estimation}, 
      author={Felix Hill and Roi Reichart and Anna Korhonen},
      year={2014},
      eprint={1408.3456},
      archivePrefix={arXiv},
      primaryClass={cs.CL},
      url={https://arxiv.org/abs/1408.3456}, 
}

@article{mikolov2013efficient,
  title={Efficient estimation of word representations in vector space},
  author={Mikolov, Tomas and Chen, Kai and Corrado, Greg and Dean, Jeffrey},
  journal={arXiv preprint arXiv:1301.3781},
  year={2013}
}

@article{frank2023,
  title={Bridging the data gap between children and large language models},
  author={Frank, Michael C},
  journal={Trends in Cognitive Sciences},
  volume={27},
  number={11},
  pages={990--992},
  year={2023},
  publisher={Elsevier}
}

@inproceedings{long2025,
  title={The {BabyView} dataset: High-resolution egocentric videos of infants' and young children's everyday experiences},
  author={Long, Bria and Sparks, Robert Z. and Xiang, Violet and Stojanov, Stefan and Yin, Zi and Keene, Grace and Tan, Alvin Wei Ming and Feng, Steven Y. and Nag, Auddithio and Zhuang, Chengxu and Marchman, Virginia A. and Yamins, Daniel L. K. and Frank, Michael C.},
  booktitle={Proceedings of the Conference on Cognitive Computational Neuroscience},
  year={2025},
  doi={10.32470/vbbjtb0}
}

@article{trueswell2013,
  title={Propose but verify: Fast mapping meets cross-situational word learning},
  author={Trueswell, John C and Medina, Tamara Nicol and Hafri, Alon and Gleitman, Lila R},
  journal={Cognitive psychology},
  volume={66},
  number={1},
  pages={126--156},
  year={2013},
  publisher={Elsevier}
}

@article{fenson1994,
  title={Variability in early communicative development},
  author={Fenson, Larry and Dale, Philip S and Reznick, J Steven and Bates, Elizabeth and Thal, Donna J and Pethick, Stephen J and Tomasello, Michael and Mervis, Carolyn B and Stiles, Joan},
  journal={Monographs of the society for research in child development},
  pages={i--185},
  year={1994},
  publisher={JSTOR}
}

@book{bloom2002,
  title={How children learn the meanings of words},
  author={Bloom, Paul},
  year={2002},
  publisher={MIT press}
}

@book{frank2021,
  title={Variability and consistency in early language learning: The Wordbank project},
  author={Frank, Michael C and Braginsky, Mika and Yurovsky, Daniel and Marchman, Virginia A},
  year={2021},
  publisher={MIT Press}
}

@article{mollica2017,
	Author = {Mollica, Francis and Piantadosi, Steven T},
	Journal = {Open Mind},
	Number = {2},
	Pages = {67--77},
	Publisher = {MIT Press},
	Title = {How data drive early word learning: A cross-linguistic waiting time analysis},
	Volume = {1},
	Year = {2017}}

@article{frank2017,
  author = {Frank, Michael C and Braginsky, Mika and Yurovsky, Daniel and Marchman, Virginia A.},
  journal = {Journal of Child Language},
  number = {3},
  pages = {677--694},
  title = {Wordbank: An open repository for developmental vocabulary data},
  volume = {44},
  year = {2017}
}

@article{smith2002,
  author  = {Smith, Linda B. and Jones, Susan S. and Landau, Barbara and Gershkoff-Stowe, Lisa and Samuelson, Larissa},
  title   = {Object Name Learning Provides On-the-Job Training for Attention},
  journal = {Psychological Science},
  volume  = {13},
  number  = {1},
  pages   = {13--19},
  year    = {2002},
  doi     = {10.1111/1467-9280.00403}
}

@article{bergelson2012,
  title={At 6--9 months, human infants know the meanings of many common nouns},
  author={Bergelson, Elika and Swingley, Daniel},
  journal={Proceedings of the National Academy of Sciences},
  volume={109},
  number={9},
  pages={3253--3258},
  year={2012},
  publisher={National Academy of Sciences}
}

@article{bergelson2017,
	author = {Bergelson, Elika and Aslin, Richard N},
	journal = {Proceedings of the National Academy of Sciences},
	number = {49},
	pages = {12916--12921},
	publisher = {National Acad Sciences},
	title = {Nature and origins of the lexicon in 6-mo-olds},
	volume = {114},
	year = {2017}}

@book{marchman2023,
	author = {Marchman, V A and Dale, P S and Fenson, L},
	publisher = {Brookes},
	title = {MacArthur-Bates Communicative Development Inventories User\'s Guide and Technical Manual, Third Edition},
	year = {2023}}

@article{bain2022whisperx,
  title={WhisperX: Time-Accurate Speech Transcription of Long-Form Audio},
  author={Bain, Max and Huh, Jaesung and Han, Tengda and Zisserman, Andrew},
  journal={INTERSPEECH 2023},
  year={2023}
}

@misc{radfordRobustSpeechRecognition2022,
  title = {Robust {{Speech Recognition}} via {{Large-Scale Weak Supervision}}},
  author = {Radford, Alec and Kim, Jong Wook and Xu, Tao and Brockman, Greg and McLeavey, Christine and Sutskever, Ilya},
  year = 2022,
  month = dec,
  number = {arXiv:2212.04356},
  eprint = {2212.04356},
  primaryclass = {cs, eess},
  publisher = {arXiv},
  doi = {10.48550/arXiv.2212.04356},
  urldate = {2023-04-20},
  archiveprefix = {arXiv}
}

\clearpage
\appendix
\setcounter{figure}{0}\renewcommand{\thefigure}{S\arabic{figure}}
\setcounter{table}{0}\renewcommand{\thetable}{S\arabic{table}}
\setcounter{equation}{0}\renewcommand{\theequation}{S\arabic{equation}}
{\centering\LARGE\bfseries Supplementary Information\par}\bigskip

\section*{Language annotation}

To filter the BabyView corpus to recordings that were primarily or exclusively in English, we sampled three 60s windows per video and used \texttt{gemini-2.5-flash} to classify the audio directly (under Vertex, a Google API that restricts the data from being used in retraining). These automatic measures correlated with child-level survey-reported proportions of English-exposure at $r=.91$. Our prompt was:

\begin{quote}\raggedright
\texttt{You identify the languages SPOKEN in short audio clips from home recordings of families with young children. Judge only from the audio. Ignore singing wordlessly, babbling, and TV/background media unless it is the only speech present.}

\texttt{Listen to this clip. Return JSON only: \{"speech": <true if any intelligible speech>, "langs": \{"<ISO 639-1>": <proportion of the SPOKEN content in that language, summing to 1>\}, "child\_directed": <true if most speech is to/with a child>\}}
\end{quote}

\section*{Reference annotation}

To annotate whether each utterance referred to an object in the matching frame, we annotated each utterance-frame pair again using \texttt{gemini-2.5-flash}. Our prompt was:

\begin{quote}\raggedright
\texttt{
Return JSON with: - "alignment": integer 0-100 for how strongly the utterance refers to a concrete object VISIBLE in the frame. 100 = clearly names a prominent, plainly-visible, central object; $\sim$50 = the named object is present but small, partial, or one of many; 0 = no visible referent (small talk, the object is absent or unidentifiable, or the utterance is not about a concrete object). Use the full range, not just the ends. - "referent": the single visible object referred to, as a lowercase common noun (e.g. "cup", "dog"); "" if alignment is low or none applies. 
}
\end{quote}

To validate these annotations, two researchers (author VAM and one other team member) manually rated a sample of 1500 frames. These frames were a stratified sample of Gemini-rated frames to ensure representation across the full rating scale from 0-100, with 50\% of frames sampled from the pool rated as unaligned. Raters made binary judgments about whether the referent was present. Human raters were provided with the same instructions as Gemini, as well as additional clarifications (e.g., that proper nouns, pronouns, and onomatopoiea should be treated as referential if their referent appears in the frame). We report precision and recall computed based on classifying Gemini ratings $\geq50$ as detections. Between-rater $\kappa$ was .75 (representing 88\% agreement) while average rater-to-Gemini $\kappa$ was .64.

\section*{DINOv3 ViT encoders}

\subsection*{Training}

We pretrained vision transformers from scratch on the BabyView data using DINOv3 self-distillation \cite{simeoni2025dinov3} via light modifications of the original implementation. We used all  9,726,507 frames extracted at 1 fps from full BabyView 2026.1 release; frames were not filtered by language content. We trained ViT-S/16 and ViT-B/16 models for 200,000 iterations at a global batch size of 512 (approximately 102M samples, or ~10.5 epochs over the corpus), and a ViT-L/16 model at a global batch size of 384, with peak learning rate set by the implementation's square-root batch-size scaling rule (effective peak $2.83 \times 10^{-3}$ at batch size 512), 16 warmup epochs, and 48 epochs of teacher-temperature warmup. Training used bfloat16 precision on a single node of 8 Nvidia A40 GPUs. 

\subsection*{Encoder probes}

To measure how well the encoders separate evaluation categories independent of learned mappings, we ran a head-free nearest-prototype probe on each encoder. Each image was represented by the mean of its cached region vectors (L2-normalized). We ran the probe on both the Konkle test set and the held out BabyView set. We computed the prototype for every word category by averaging the representation of each exemplar in the test set. For each 4AFC trial, we asked whether the cosine similarity of the target image to the prototype of its category was higher than to the prototype of any of the three foil categories. 

\section*{Language data processing and model training}

Architectural details follow \cite{vong2024grounded}. Prior to training contrastive models, utterance data were tokenized and normalized. Each utterance was converted to lower case and split by whitespace; non-alphabetic characters (including punctuation and digits) were removed. Model vocabulary was built separately for every training run from all word types occurring more than 5 times in the training corpus (15,608 word types in the full corpus). Types with fewer than 5 instances were deleted; since there is no representation of word order this did not cause any issues in sequence estimation. Utterances were truncated to a maximum of 16 in-vocabulary tokens. 

During training, each retained token is mapped to a learned 512-dimensional embedding; utterance vectors are the L2-normalized mean of these embeddings. The similarity score between an utterance and a frame is the maximum over frame regions of the cosine similarity between the region's projected vector and the utterance vector. The final score is scaled by a learned temperature (initialized at 1/0.07, capped at 100) and used in a symmetric InfoNCE objective over batches of 256 pairs, averaging the image-to-text and text-to-image cross-entropy losses. 

\section*{Evaluation details}

\subsection*{Konkle evaluation}

Our primary evaluation followed \cite{vong2024grounded} in using 60 categories from Konkle et al.'s white-background categorization stimuli \cite{konkle2010conceptual}. We created a 117-category disjoint development split for purposes of selecting the training epoch. On each 4AFC trial, we drew one image at random from the category pool for the target and each distractor. We then scored each candidate image against the target word's embedding by taking the maximum similarity between the word and the frame over the 16-region MIL grid. We ran 100 trials per category to remove category exemplar noise and report the mean over categories. Out-of-vocabulary categories were scored at chance. Categories with fewer than four images were excluded. 

\subsection*{In-domain evaluation}

 We created a complementary in-domain word learning benchmark using the 1,258 videos excluded by the audio-language filter and their Gemini referent annotations. Candidate frames had a maximal referential alignment rating. We excluded person terms, body parts, and generic containers because all were highly frequent (e.g., ``hand'' occurs in a large proportion of frames). We included at most one frame per video per category. As above, we included a category if we could identify at least four exemplars. The final set of images contained 603 frames in 82 categories from 356 videos. Scoring was as above for the Konkle evaluation. 

\subsection*{LEVANTE-bench evaluation}

We used the 4AFC vocabulary evaluation from LEVANTE-bench \cite{tan2026levante}. This is a 159-item set, with a single fixed photo for each item and distractor. Photos vary but often consist of an object in context. Models were scored as per the Konkle evaluation above, with additional scoring on in-vocabulary items only shown in Figure \ref{fig:levante} below. We downloaded data from \texttt{levante-data-pilots} (\url{https://doi.org/10.57761/pare-hq93}). Because the LEVANTE vocabulary measure was adaptive and children see trials of different difficulties depending on their estimated ability, average accuracies are not informative about developmental change. Thus, children's average accuracy across ages was computed as in the original paper, by using model-estimated child-level ability scores from the LEVANTE \texttt{scores} table and reconstructing expected full-instrument accuracies, then averaging. We used only data from English-speaking children. Age groups with fewer than 8 children were excluded. Confidence intervals were computed via a bootstrap over runs. The final sample included 145 English-speaking children, and 142 of 159 items were present in the BabyView transcripts.

\subsection*{Sigmoid curve fitting}

We fit scaling curves using a three-parameter logistic over $\log_{10}$ training examples,
\begin{equation*}
  \mathrm{acc}(N) \;=\; c \;+\; \frac{A - c}{1 + \exp\!\big(-(\log_{10} N - m)/s\big)}.
\end{equation*}
\noindent We fixed the floor $c$ at 25\% and fit the upper asymptote $A$, the midpoint $m$, and the slope $s$. Fits were computed via least squares weighted by across-seed standard deviations. We computed uncertainty bands using Monte Carlo draws computed by perturbing each point proportional to its standard error. 95\% confidence bands were computed empirically from 4000 of these draws. 

\section*{Developmental data}

\subsection*{Conversion of Wordbank data to Konkle scores}

To estimate young children's Konkle score, we estimated possible scores from parent-reported data in Wordbank \cite{frank2017}. Wordbank archives parent reports using the MacArthur-Bates Communicative Development Inventory \cite{fenson1994,marchman2023}. Parents of younger children (Words \& Gestures form; 8 -- 18 months) were asked whether their child produces or understands each word; older children (Words \& Sentences form; 16 -- 30 months) are asked only about production. We downloaded all American English CDI administrations in the target ages ($N=2807$ administrations for WG and $N=5721$ for WS) and matched the 60 Konkle categories to CDI words: 40/60 words matched on the WG form and 46/60 matched on the WS. Unmatched words included lower-frequency words (e.g., \emph{guitar}, \emph{umbrella}, \emph{trumpet}, \emph{rug}). Because the unmatched words are also harder for the models, we score both children (WS and WG) and models on the same 40 WG-matched words. For each age group, we computed the proportion of children reported to produce or understand the target word. We then converted these proportions to 4AFC scores by assuming that the child responds correctly if they know the word and guesses otherwise. 

\subsection*{Developmental time conversion}

We converted frame--utterance pairs to years of input by assuming 4,000 waking hours per year ($\sim11$ hours per day) and 697 utterances per hour. The waking hour conversion is approximate, given that children's total waking time varies over their first years of life and there is substantial individual variation in total waking time. We estimated speech rate based on our own data (per-child median 702, 95\% CI 637--747). We expect that in practice this number may be an upper bound because parents are likely to record during parts of the day that contain more talk. 

\section*{Analysis of learned lexicons}

To identify nouns included in Figure \ref{fig:lexicon}A, we filtered nouns with frequency $\geq50$. Many lexical items had representations that did not move far from random initialization, so we included only nouns whose top 3 nearest neighbors had cosine distance that was $>4\times$ the distance between random vectors. tSNE maps were constructed with perplexity parameter 30. For Figure \ref{fig:lexicon}B, we constructed null distributions by permuting category labels 5,000 times for each of three seeds. For Figure \ref{fig:lexicon}C, the word-relatedness benchmark uses five different similarity judgment datasets (WordSim-353, SimLex-999, RG-65, MTest-3000, SimVerb-3500). We $z$-scored within each set. We restricted evaluation to words present in the training data at each scale separately.

To estimate similarity structures due to word-word co-occurrence, we trained a word2vec baseline on the same set of utterances used in contrastive VLM training. We fit a skip-gram model with negative sampling (gensim 4.4.0; 300 dimensions, window 5, 10 negative samples, 10 epochs, minimum count 5, fixed seed, single worker for determinism). Each utterance was treated as a separate sentence. We trained a separate word2vec baseline for each training-set size. At evaluation, we L2-normalized word vectors and computed cosine similarities.

\begin{table*}[ht!]\centering\footnotesize\setlength{\tabcolsep}{3pt}
\caption{Training runs underlying all figures. Every run uses a frozen encoder, region-MIL head (unless noted), InfoNCE, 20 epochs, and dev-117 epoch selection; quantities are designed training pairs (full corpus = 1,686,105).}
\label{tab:experiments}
\begin{tabular}{p{3.1cm}p{1.5cm}p{4cm}p{0.8cm}p{0.7cm}p{2.3cm}p{2.4cm}}\toprule
Experiment & Encoders & Training pairs & Seeds & Runs & Evaluation & Figures \\\midrule
Scaling (random subsamples) + full corpus & all six & 3k; 10k; 30k; 100k; 300k; 1M; full (1.69M) & 3--8 & 270 & Konkle; LEVANTE; in-domain; lexicon & \ref{fig:scaling}, \ref{fig:lexicon}, \ref{fig:development}, \ref{fig:encoder_probe}, \ref{fig:indomain}, \ref{fig:item_difficulty}, \ref{fig:levante}, \ref{fig:lexicon_alignment}, \ref{fig:lexicon_relatedness}, \ref{fig:lexicon_tsne}, \ref{fig:nomil}, \ref{fig:window} \\
Aligned-pair scaling & all six & 10k; 30k; 100k; 172k & 3 & 72 & Konkle; lexicon & \ref{fig:scaling}, \ref{fig:alignment_encoders}, \ref{fig:lexicon_alignment} \\
Diversity (children at fixed budget) & OTS-304M & 30k budget; 100k budget; 300k budget & 5 & 65 & Konkle & \ref{fig:diversity} \\
No-MIL (whole-frame) control & all six & 30k; 300k; full (1.69M) & 5 & 90 & Konkle & \ref{fig:nomil} \\
Alignment-selection controls & all six & 172k (plain random); 172k (matched, unaligned); 1.51M (full$-$matched); 172k (aligned); 1.51M (full$-$aligned); 1.51M (full$-$random) & 3 & 108 & Konkle & \ref{fig:alignment_controls} \\
Child-speech (KCHI) control & OTS-304M & 1.30M (no child speech); 1.30M (matched) & 3 & 6 & Konkle & \ref{fig:kchi} \\
Temporal-window ($\pm$5 s) control & all six & 30k; 300k; 1M; full (1.69M) & 3--5 & 84 & Konkle & \ref{fig:window} \\
\bottomrule\end{tabular}\end{table*}

\begin{figure}
\centering
\includegraphics[width=\textwidth]{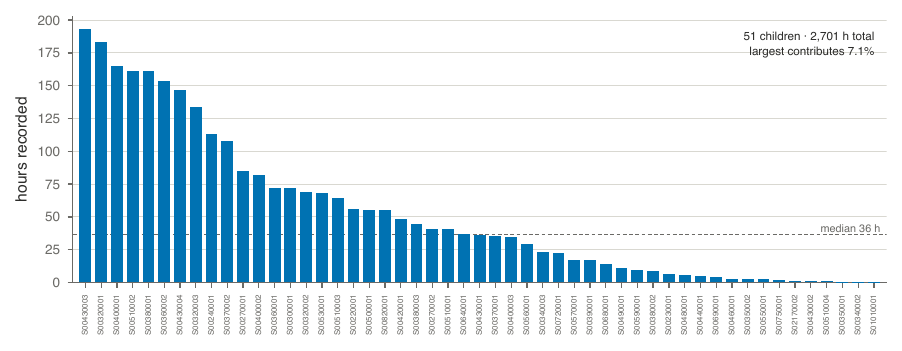}
\caption{Histogram of the number of hours recorded by each child included in the full 2026.1 BabyView release.}
\label{fig:corpus_effort}
\end{figure}

\begin{figure}
\centering
\includegraphics[width=\textwidth]{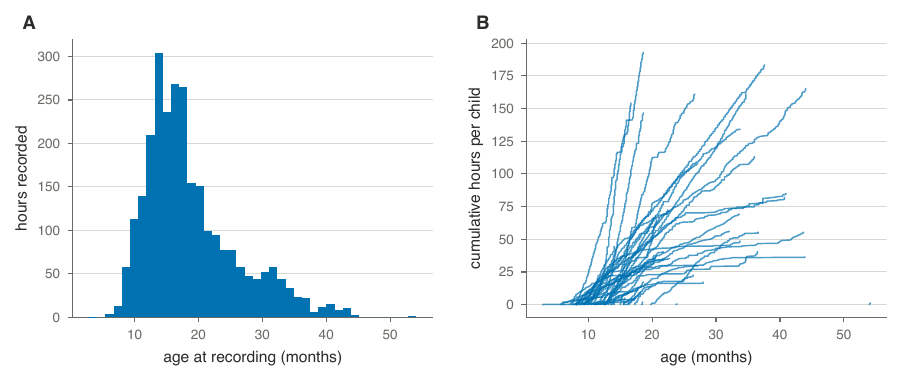}
\caption{(A) Histogram of the target child's age at the time of each recording. (B) Cumulative hours of recording per child for each child in the corpus.}
\label{fig:corpus_age}
\end{figure}

\begin{figure}
\centering
\includegraphics[width=\textwidth]{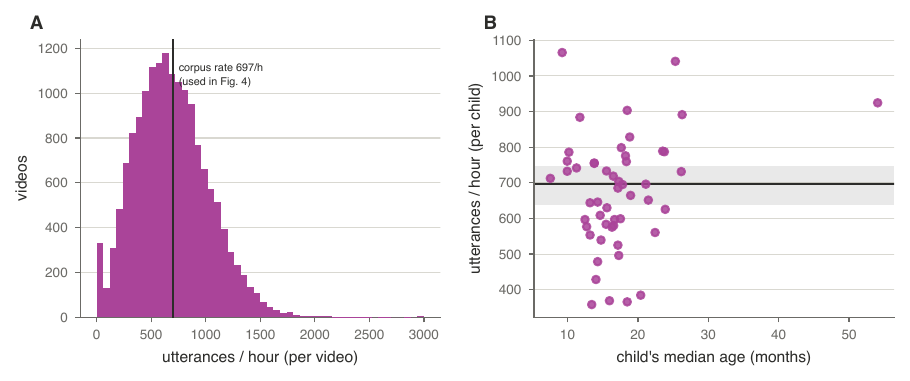}
\caption{(A) Histogram of total utterances per hour. (B) Utterances per hour plotted by child's median age.}
\label{fig:speech_density}
\end{figure}

\begin{figure}
\centering
\includegraphics[width=\textwidth]{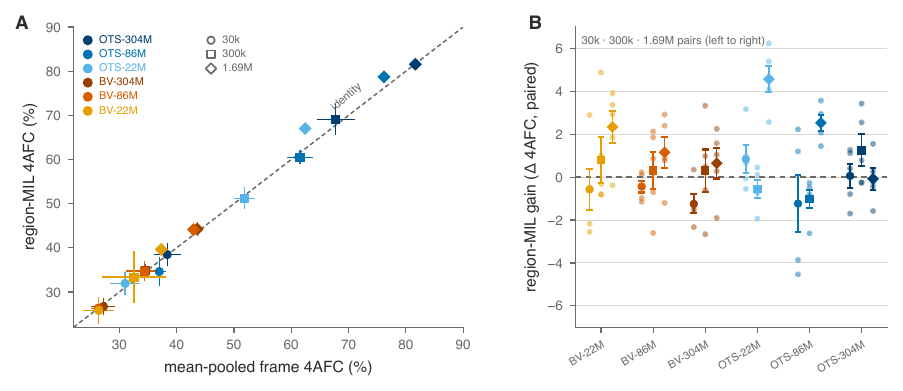}
\caption{Comparison of models with and without multiple-instance learning (MIL) across sub-regions of the frame. (A) Mean-pooled vs. MIL accuracies for all encoders and three training data scales. (B) Difference scores for each model, scale and seed. }
\label{fig:nomil}
\end{figure}

\begin{figure}
\centering
\includegraphics[width=\textwidth]{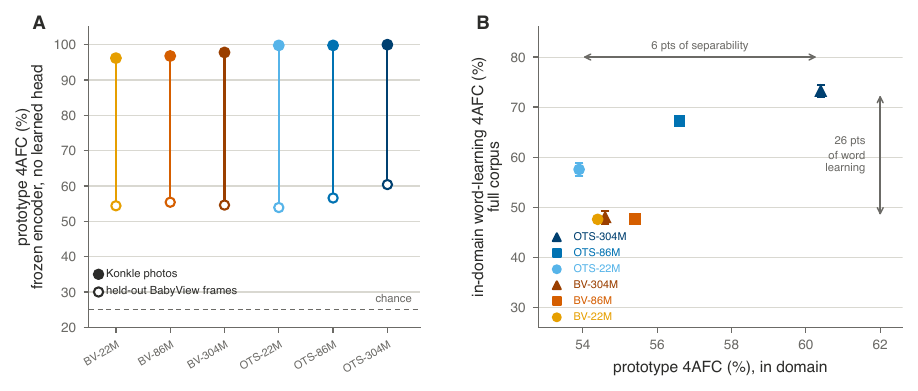}
\caption{(A) Similarity of target images to their category prototype for each test set (Konkle, held-out BabyView frames) and encoder; similarity is measured by accuracy in categorization judgments in a 4AFC. (B) Word learning performance for the full corpus plotted by prototype 4AFC.}
\label{fig:encoder_probe}
\end{figure}

\begin{figure}
\centering
\includegraphics[width=\textwidth]{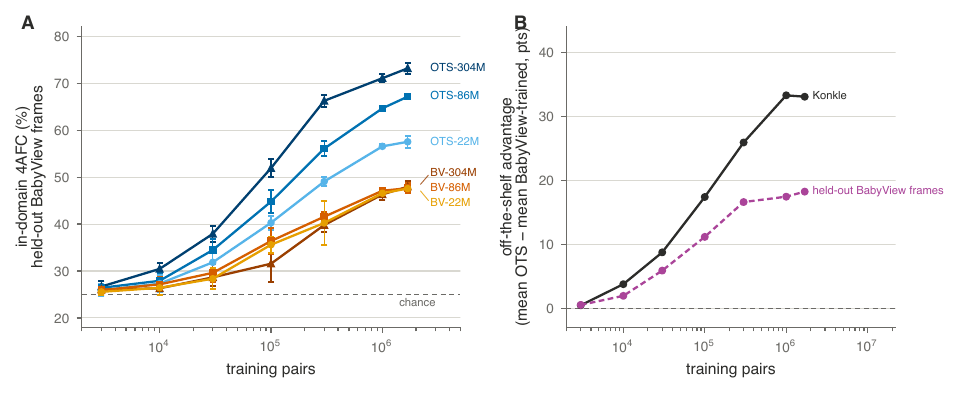}
\caption{In-domain evaluation on held-out BabyView frames (from non-English videos). (A) Accuracy for each encoder plotted by number of training pairs. Error bars show between-seed standard deviation. (B) Advantage for OTS vs. BV-trained models (average across model sizes) for both Konkle and held-out BabyView frames.}
\label{fig:indomain}
\end{figure}

\begin{figure}
\centering
\includegraphics[width=\textwidth]{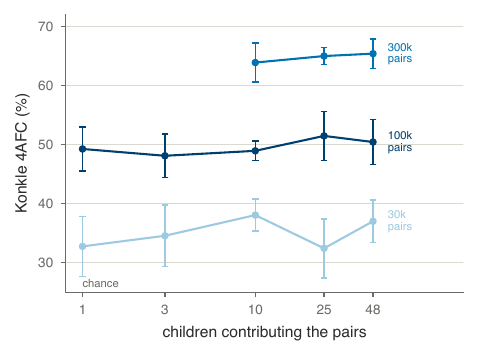}
\caption{Konkle evaluation performance by dataset diversity. Results come from OTS-304M runs with matched total number of frame-utterance pairs, but with different numbers of children's data contributing. Error bars show between-seed standard deviation.}
\label{fig:diversity}
\end{figure}

\begin{figure}
\centering
\includegraphics[width=\textwidth]{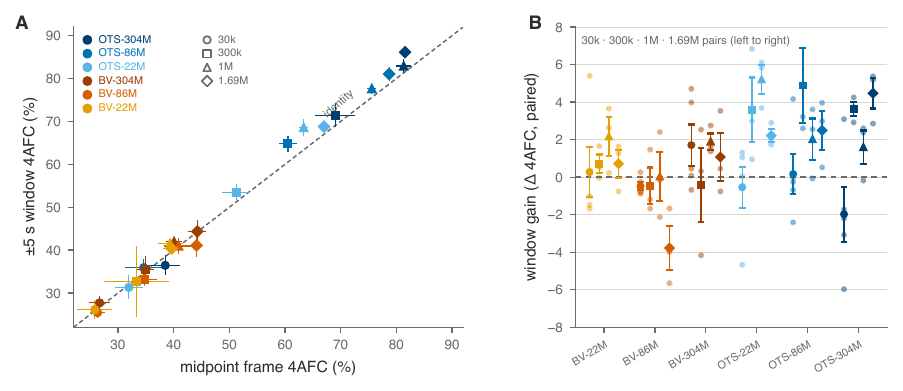}
\caption{Comparison of models with and without multiple-instance learning (MIL) across temporal windows ($\pm5$s, for a total of 11 frames). (A) Midpoint frame vs. MIL Konkle accuracies for all encoders and four training data scales. (B) Difference scores for each model, scale and seed.}
\label{fig:window}
\end{figure}

\begin{figure}
\centering
\includegraphics[width=\textwidth]{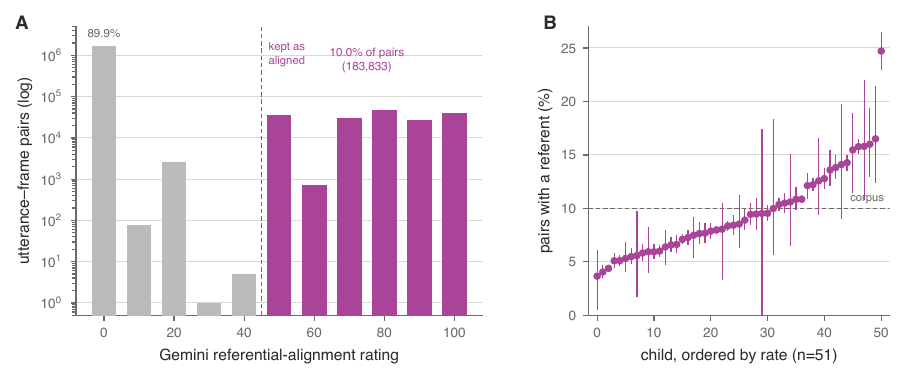}
\caption{(A) Log count histogram of Gemini referential alignment ratings, with retained pairs shown in purple. (B) Proportion aligned pairs by child, ordered by rate.}
\label{fig:alignment_scores}
\end{figure}

\begin{figure}
\centering
\includegraphics[width=\textwidth]{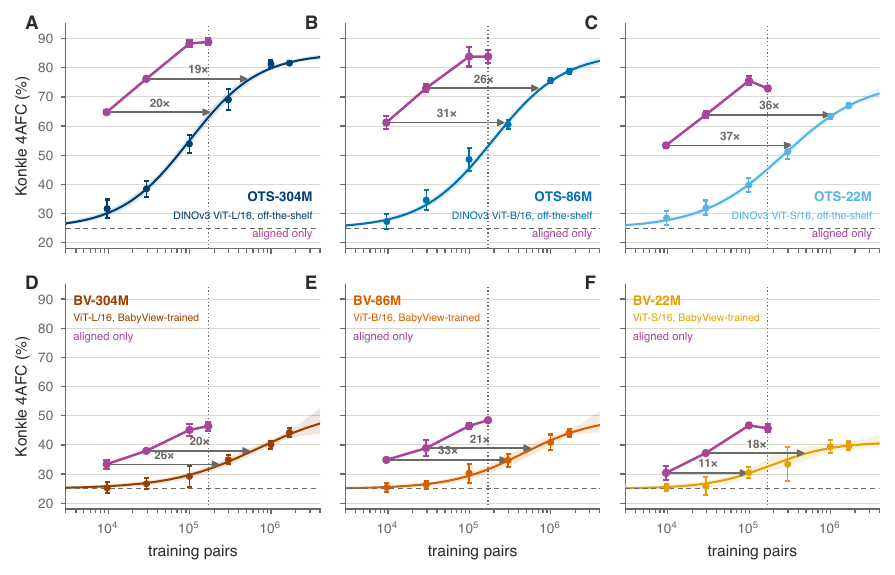}
\caption{Full results for aligned data across all encoders. Plotting conventions are as in Figure \ref{fig:scaling}.}
\label{fig:alignment_encoders}
\end{figure}

\begin{figure}
\centering
\includegraphics[width=\textwidth]{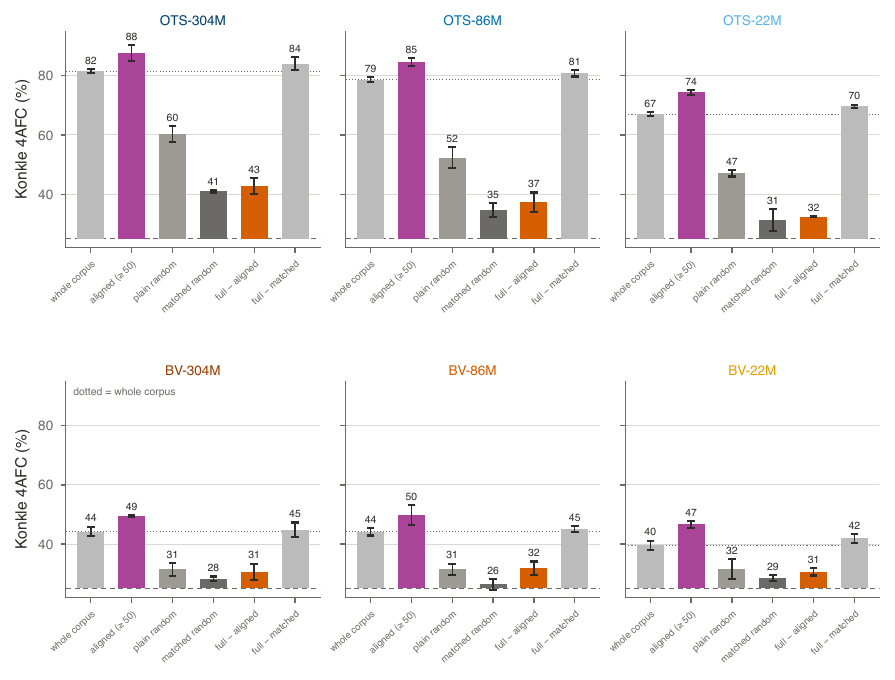}
\caption{Accuracy for each encoder, plotted by data subsample. Whole corpus accuracy is compared to: accuracy on the aligned subsample alone; a size-matched random subsample; a size-matched subsample of unaligned pairs, matched to the aligned set's joint distribution of target-noun and utterance-length; the full corpus with the aligned subsample removed; and the full corpus with the matched unaligned subsample removed. Aligned scaling arms train on the N highest-rated pairs; the alignment-selection controls train on the full set.}
\label{fig:alignment_controls}
\end{figure}

\begin{figure}
\centering
\includegraphics[width=\textwidth]{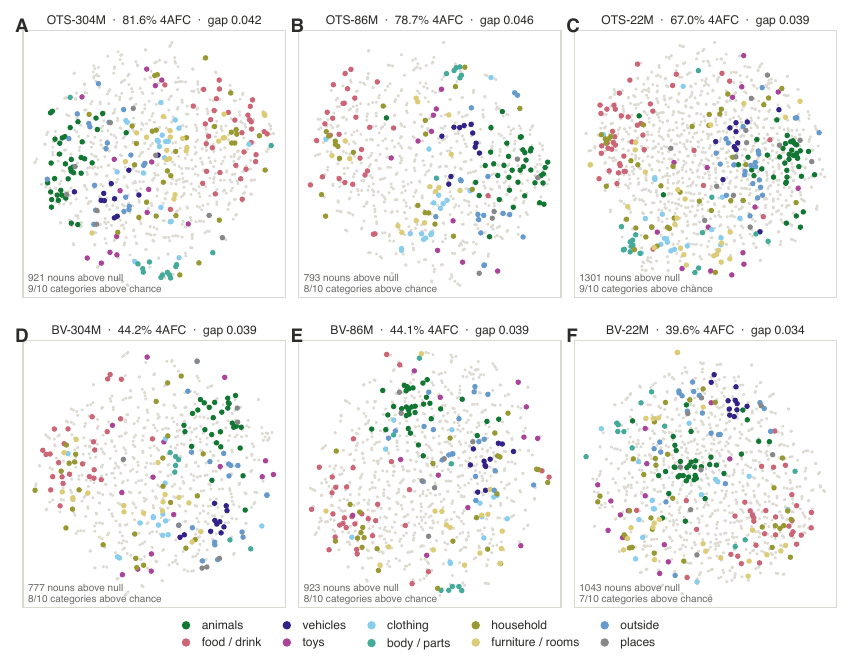}
\caption{tSNE results for all six encoders. Plotting conventions are as in Figure \ref{fig:lexicon}A.}
\label{fig:lexicon_tsne}
\end{figure}

\begin{figure}
\centering
\includegraphics[width=\textwidth]{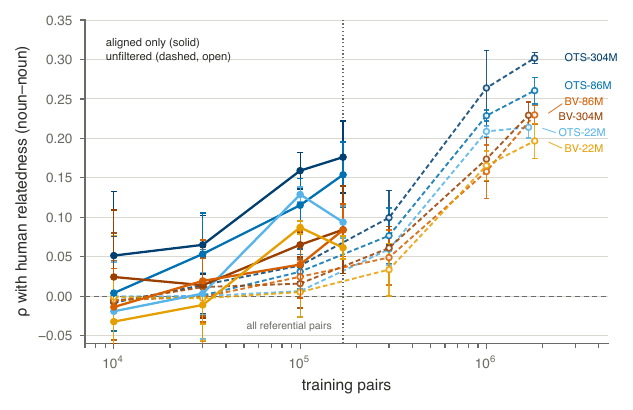}
\caption{Spearman correlation with human relatedness judgments (nouns only) from models trained on only referentially-aligned pairs (solid lines) compared with all pairs (dashed lines).}
\label{fig:lexicon_alignment}
\end{figure}

\begin{figure}
\centering
\includegraphics[width=\textwidth]{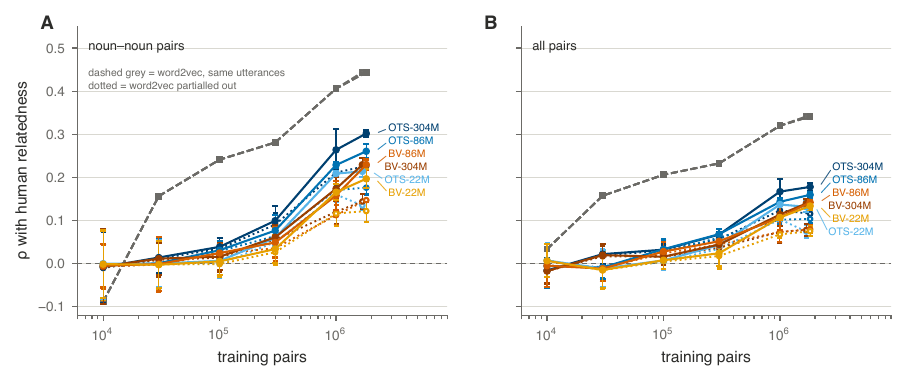}
\caption{(A) Spearman correlation with human relatedness judgments for noun-noun pairs; results for all encoders. (B) Results across all words. Plotting conventions are as in Figure \ref{fig:lexicon}C.}
\label{fig:lexicon_relatedness}
\end{figure}

\begin{figure}
\centering
\includegraphics[width=\textwidth]{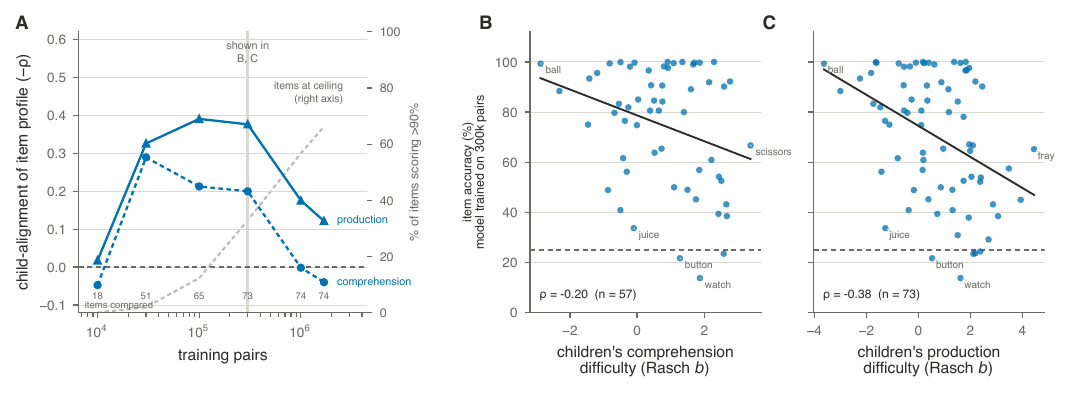}
\caption{(A) Alignment (negative correlation) between children's item difficulty and OTS-304M model item accuracy, across model training scales. Alignments are computed on the 74/177 Konkle items in the test and dev set with a corresponding MB-CDI item, for production and comprehension. Child difficulties are derived from a Rasch model. Because model accuracy reaches ceiling, correlations decline at larger scales. Numbers show the total number of items in the model vocabulary at each scale. The gray dashed line shows the proportion of items with model accuracy $>90\%$. (B) Model accuracy correlation with comprehension difficulty at 300k training pairs. (C) Model accuracy correlation with production difficulty at 300k training pairs.}
\label{fig:item_difficulty}
\end{figure}

\begin{figure}
\centering
\includegraphics[width=\textwidth]{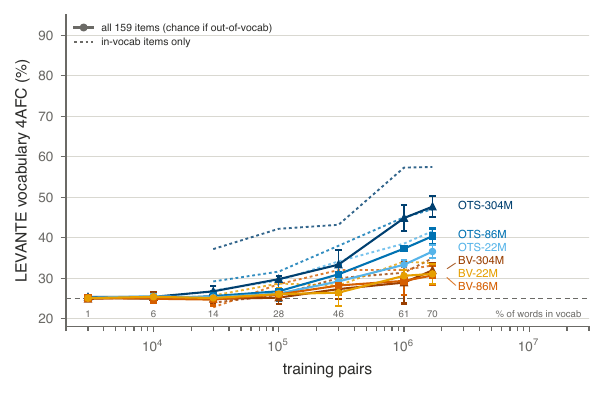}
\caption{Full results on the LEVANTE-bench vocabulary assessment for all embedding models. Solid lines show results across all items; dashed lines show results from only those items that were in the models' vocabulary for each training dataset size.}
\label{fig:levante}
\end{figure}

\begin{figure}
\centering
\includegraphics[width=\textwidth]{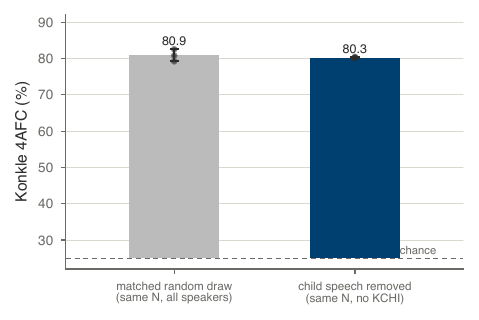}
\caption{Comparison of accuracy for the OTS-304M model on the corpus with key-child utterances removed (blue) and a matched random draw (gray).}
\label{fig:kchi}
\end{figure}

\newpage

\end{document}